\documentclass{llncs}
\usepackage{times,mathptm,amsmath,epsfig,graphicx}     
\usepackage[T1]{fontenc}
\usepackage[cp1250]{inputenc}

\newcommand\R{$\cal R$}

\newcommand\be{\begin{equation}}
\newcommand\ee{\end{equation}}
\newcommand\bea{\begin{eqnarray}}
\newcommand\eea{\end{eqnarray}}

\def\bW{{\mathbf W}} \def\bS{{\mathbf S}}
\def\bX{{\mathbf X}} \def\bY{{\mathbf Y}} \def\bR{{\mathbf R}}

\begin{document}

\title{Neural eliminators and classifiers}

\author{W{\l}odzis{\l}aw Duch\inst{1}, Rafa{\l} Adamczak\inst{1} and Yoichi Hayashi\inst{2}
\thanks{Partially supported by the Polish Committee of Scientific Research, grant no. 8 T11C 006 19. First version of this paper was presented at the 7th International Conference on Neural Information Processing 2000 (ICONIP), Dae-jong, Korea, ed. by Soo-Young Lee, pp. 1029 - 1034.}}
\institute{Dept. of Informatics, Nicholaus Copernicus University, Grudzi\c{a}dzka 5, Toru\'n, Poland.  http://www.is.umk.pl/kmk\\
\and
Department of Computer Science, Meiji University, Kawasaki 214-8571, Japan. 
   Email: hayashiy@cs.meiji.ac.jp} 

\maketitle


\begin{abstract}

Classification may not be reliable for several reasons: noise in the data, insufficient input information, overlapping distributions and sharp definition of classes. Faced with several possibilities neural network may in such cases still be useful if instead of a classification elimination of improbable classes is done. Eliminators may be constructed using classifiers assigning new cases to a pool of several classes instead of just one winning class. Elimination may be done with the help of several classifiers using modified error functions. A real life medical application of neural network is presented illustrating the usefulness of elimination.

\end{abstract}

\begin{keywords}
Classification, elimination of classes, neural networks, decision support, medical diagnosis, error functions.
\end{keywords}

\section{Introduction.}

Neural, fuzzy and machine learning systems are usually applied as classifiers or approximators. In real-world problems designation of classes may be problematic due to the approximate nature of linguistic concepts labeling cases that change in a continuous way. For example, medical databases contain names of diseases that may develop in time, from mild to severe cases, with intermediate or mixed forms. Corresponding class distributions $p(\bX|C_i)$ will strongly overlap requring fuzzy class labels. The information provided in the database may be insufficient to distinguish the classes although they may be separable by some unknown features (for example, results of new medical test). In such situations reliable classification is not possible and comparison of results based on the number of classification errors may be quite misleading.

If soft class labels are needed or if insufficient number of classes is defined some conclusions can still be drawn by looking at the classification probabilities. For example, the system may assign the new case given for evaluation to the overlapping region where two or more classification probabilities have significant values, in a way creating new, mixed or border classes. Introduction of new classes cannot be done automatically and requires close collaboration with domain experts.
An alternative way of solving such problems is to eliminate improbable classes, predicting that the unknown case belongs to a subset of $k$ classes out of $K$ possible ones. To account for the possibility of class distributions overlapping in a different way in different regions of the input space the number $k$ should not be fixed. Such systems may be called {\bf eliminators} since their primary goal is to eliminate with high confidence classes that are improbable.

Any model $M$ that estimates probabilities of classification $p(C_i|\bX;M)$ may be used to create new, soft class labels or to eliminate some classes predicting that $\bX$ belongs to two or more classes. In particular neural and neurofuzzy systems are well suited for this purpose, although they should be modified to optimize elimination of several classes rather then the prediction of a single class. Some other classification systems, such as statistical discrimination methods, support vector machines \cite{SVM}, decision trees or the nearest neighbor methods provide only sharp yes/no classification decisions \cite{Statlog}. Detailed interpretation of a given case is possible if methods of explanatory data analysis displaying the new case in relation to the cases stored in the training database are used, or if classification confidence intervals are calculated \cite{dataunderst}.

Our goal in this paper is twofold. In the next section problems specific to class elimination in neural networks are disussed, followed by a presentation of a universal method for estimation of probabilities that is applicable to any classifier. A real-life example of a difficult medical problem is presented in the fourth section and a short discussion concludes this paper.

\section{Elimination instead of prediction}

Consider a classification problem in $N$ dimensions with two overlapping classes described by Gaussian distributions with equal covariance matrices $\Sigma$:

\[ \nonumber 
  p(\bX|C_i)=\frac{1}{(2\pi)^{\frac{N}{2}}|\Sigma|^{\frac{1}{2}}}
  \exp\left\{-\frac{1}{2}(\bX-\bar\bX_i)^T \Sigma^{-1} (\bX-\bar\bX_i)\right\} \nonumber
\]

Using Bayes' theorem the posterior probability for the first class is \cite{bishop}:

\be \label{eq:pprob}
p(C_1|\bX)=\frac{p(\bX|C_1)P(C_1)}{p(\bX|C_1)P(C_1) + p(\bX|C_2)P(C_2)}
\ee
The $P(C_k)$ are {\em a priori} class probabilities. Thus $p(C_1|\bX)= \sigma(y(\bX))$, where the function $y(\bX)$ is:

\be \label{eq:ylog}
y(\bX)=\ln\frac{p(\bX|C_1)P(C_1)}{p(\bX|C_2)P(C_2)} = \bW\cdot\bX-\theta
\ee
where

\be
\bW= (\bX_2-\bX_1)^T \Sigma^{-1} = \bW\cdot\bX-\theta
\ee
and $\theta=\theta(\bX_1,\bX_2,\Sigma,P(C_1),P(C_2))$.
The posterior probability is thus given by a specific logistic output function. For more than two classes normalized exponential functions (called also softmax functions) are obtained by the same reasoning:

\be \label{eq:softm}
p(C_k|\bX)=\frac{\exp(y_k(\bX))}{\sum_i \exp(y_i(\bX))}
\ee
These normalized exponential functions may be interpreted as probabilities. They are provided in a natural way by multilayer perceptron networks (MLPs). If one of the probabilities is close to 1 the situation is clear. Otherwise $\bX$ belongs to the border area and a unique classification may not be possible. The domain expert should decide if it makes sense to introduce a new, mixed class, or to acknowledge that insufficient information is available for accurate classification.

\subsection{Measures of classifier performance}

Measures of classifier performance based on accuracy of confusion matrices $F(C_i,C_j)$ do not allow to evaluate their usefulness. Introduction of risk matrices or use of receiver-operator characteristic (ROC) curves \cite{ROC} does not solve the problem either.

If the standard approach fails to provide sufficiently accurate results for some classes one should either attempt to create new classes or to minimize the number of errors between a temporary new class composed of two or more distinct classes. This requires a modification of the standard cost function.
Let $C(\bX)$ be the true class of the vector $\bX$ and $p(C|\bX;M)$ the probability of class $C$ calculated using the model $M$.
The neural cost function should minimizes the error:

\be
E_2(\{\bX\},R;M)= \sum_i \sum_{\bX} H\left(p(C_i|\bX)-\delta(C_i,C(\bX))\right)   \nonumber
\ee
where $i$ runs over all different classes and $\bX$ over all training vectors,
$C(\bX)$ is the true class of the vector $\bX$ and the function $H(\cdot)$ should be monotonic and positive; most often the quadratic function or the entropy-based function is used. $M$ specifies all adaptive parameters and variable procedures of the classification model that may affect the cost function.

Risk matrix of the overall classification $R(C_i,C(\bX))$ may easily be included in this cost function.
The elements of the risk matrix $R(C_i,C_j)$ are proportional to the risk of assigning the $C_i$ class when the true class is $C_j$. In the simplest case $R(C_i,C_j)=1-\delta_{ij}$. Regularization terms aimed at minimization of the complexity of the classification model are frequently added to the cost function, allowing to avoid the overfitting problems. To improve generalization the sum should run over all training examples $\bX$ but the model $M$ used to compute $p(C_i|\bX)$ should be created without the $\bX$ vector in the training set.

Another form of the cost function is also useful:

\bea
C_j(\bX^p) \leftarrow j &=& \arg \max_i\ p(C_i|\bX^p;M) \\
E(\left\{ X \right\};M) &=&\sum_p K\left(C(\bX^p)-C_j(\bX^p)\right) \nonumber
\eea
where $C_j(\bX^p)$ corresponds to the best recommendation of the classifier and the kernel function $K(\cdot,\cdot)$ measures similarity of the classes. A general expression is:

\be
E\left( {\left\{ X \right\};M} \right)=
\sum\limits_i {K\left( {d\left( {X^{(i)},R} \right)} \right)}Err\left( {X^{(i)}} \right)
\ee

For example in the local regression based on the minimal distance approaches \cite{localw} the error function is:

\be
E(\left\{ X \right\};M)=\sum_{p} K(D(\bX^p,\bX^{ref})) (F(\bX^p;M)-y^p)^2
\ee
where $y^i$ are the desired values for $\bX^i$ and $F(\bX^i;M)$ are the values predicted by the model $M$. Here the kernel fun\-ction $K(d)$ measures the influence of the reference vectors on the total error. If $K(d)$ has a sharp high peak around $d=0$ the function $F(\bX;M)$ will fit the values corresponding to the reference input vectors almost exactly and will make large errors for other values. In classification problems kernel function will determine the size of the neighborhood (around the known cases) in which accurate classification is required.

Suppose that both off-diagonal elements $F_{12}$ and $F_{21}$ of the confusion matrix are large, i.e. that the first two classes are frequently mixed. These two classes may be separated from all the others using an independent classifier. The joint class is designated as $C_{1,2}$ and the model trained with the following cost function:

\bea \label{err2}
&&E_d(\left\{ X \right\};M) = \sum_{\bX} H\left( p(C_{1,2}|\bX;M)-\delta(C_{1,2},C(\bX))\right)   \nonumber\\
&&+\sum_{k>2} \sum_{\bX} H\left(p(C_k|\bX;M)-\delta(C_k,C(\bX))\right)
\eea
where $\delta(C_{1,2},C(\bX))$ is 1 if $C(\bX)$ is 1 or 2.

Training with such error function provides new, possibly simpler, decision borders. In practice one should use classifier first and only if classification is not sufficiently reliable (several probabilities are almost equal) try to eliminate subsets of classes. If joining pairs of classes is not sufficient triples and higher combinations may be considered.

In the image analysis community two coefficients, $\kappa$ and $\tau$, are commonly used to measure  classifier's performance. The $\kappa$ coefficient \cite{kappa} corrects the accuracy for chance agreement and is calculated as:

\begin{equation}
\kappa = \frac{N \sum_{i=1}^c F_{ii} - \sum_{i=1}^r (F_{i+} F_{+i})}
{N^2 - \sum_{i=1}^c (F_{i+} F_{+i})},
\end{equation}

where $N$ is the number of classified cases, $c$ is the number of classes (including the ``unknown" or rejected class, i.e. $c$ is the number of rows in the confusion matrix), $F_{ii}$ is the number of cases correrctly assigned to the class $C_i$, $x_{i+}$ is the row sum for row i and $x_{+i}$ is the column sum for column $i$.
The Tau coefficient \cite{tau} is calculated by:

\be
p_0=\frac{1}{N} \sum_{i=1}^c F_{ii};\quad
\tau =\frac{p_0-p_r}{1-p_r},
\ee
where $p_0$ is the overall accuracy, $F_{ij}=F(C_i,C_j)$ is the confusion matrix element, and $p_r$ is the base rate (maximum {\em a-priori} probability of a class membership). This coefficent is zero for prediction accuracies equal to the base rate, negative if these predictions are below the base rate and reaches one for perfect predictions. Confidence intervals for $p_0$ and $\tau$ may be taken as \cite{tau}:

\bea
\sigma ^2(p_0) &=& \frac{1}{N}p_0(1-p_0); \\
\sigma ^2(\tau) &=& \frac{\sigma ^2(p_0)}{(1-p_r)}.
\eea

Then comparing two results the Z-score:

\begin{equation}
Z=\frac{\tau_1-\tau_2}{\sqrt{\sigma^2(\tau_1)+\sigma^2(\tau_2)}},
\end{equation}

for statistically significant differences between these results at the 95 \% confidence level corresponds to $\left| Z\right| \geq 1.96$.

Although these coefficients are useful and should be used instead of quoting accuracy the problem lies in creating new classes and eliminating other classes when reliable classification is not possible. An approach to image analysis in which arbitrarily created class names are joint together has been described \cite{Dutra}. It is based on evaluation of class grouping using the Jeffreys-Matushita distance \cite{Fukunaga} for evaluation of separation of two distributions. Although this may be a useful approach in remote sensing applications in other applications the problem lies not in joining the whole classes but rather recognizing the border cases.

\section{Calculation of probabilities}

Some classifiers do not provide probabilities, therefore it is not clear how to optimized them for elimination of classes instead of selection of the most probable class. A universal solution independent of any classifier system is described below.

Real input values $\bX$ are obtained by measurements that are carried with finite precision. The brain uses not only large receptive fields for categorization, but also small receptive fields to extract feature values. Instead of a crisp number $X$ a Gaussian distribution $G_{X}=G(Y;X,S_{X})$ centered around $X$ with dispersion $S_{X}$ should be used.
Probabilities $p(C_i|\bX;M)$ may be computed for any classification model $M$ by performing a Monte Carlo sampling from the joint Gaussian distribution for all continuous features $G_{\bX}=G(\bY;\bX,\bS_X)$.  Dispersions  $\bS_X = (s(X_1), s(X_2)\dots s(X_N))$ define the volume of the input space around $\bX$ that has an influence on computed probabilities. One way to ``explore the neighborhood'' of $\bX$ and see the probabilities of alternative classes is to increase the fuzziness $\bS_X$ defining $s(X_i) = (X_{i,max}-X_{i,min}) \rho$, where the parameter $\rho$ defines a percentage of fuzziness relatively to the range of $X_i$ values.

With increasing $\rho$ values the probabilities $p(C_i|\bX;\rho, M)$ change. For sufficiently large $\rho$ the {\em a priori} class probabilities should be recovered. Even if a crisp rule-based classifier is used non-zero probabilities of classes alternative to the winning class will gradually appear. The way in which these probabilities change shows how reliable is the classification and what are the alternatives worth remembering. If the probability $p(C_i|\bX;\rho, M)$ changes rapidly around some value $\rho_0$ the case $\bX$ is near classification border and an analysis of $p(C_i|\bX;\rho_0, s_i, M)$ as a function of each $s_i=s(X_i), i=1\dots N$ is needed to see which features have strong influence on classification. Displaying such probabilities allows for precise evaluation of the new data also in cases where analysis of rules is too complicated. A more detailed analysis of these probabilities based on \emph{confidence intervals} and \emph{probabilistic confidence intervals} has recently been presented by Jankowski \cite{iis00}. Confidence intervals are calculated individually for a given input vector while logical rules are extracted for the whole \emph{training set}.

Confidence intervals measure maximal deviation from the given feature value $X_i$ (assuming that other features of the vector $\bX$ are fixed) that do not change the most probable classification of the vector $\bX$.
If this vector lies near the class border the confidence intervals are narrow, while for vectors that are typical for their class confidence intervals should be wide.
These intervals facilitate precise interpretation and allow to analyze the stability of sets of rules.

For some classification models probabilities $p(C_i|\bX;\rho, M)$ may be calculated analytically. For the crisp rule classifiers \cite{duch_iis99} a rule $R_{[a,b]}(X)$, which is true if $X\in [a,b]$ and false otherwise, is fulfilled by a Gaussian number $G_X$ with probability:

\be
p(R_{[a,b]}(G_X)=T)\approx \sigma(\beta(X-a))-\sigma(\beta(X-b))
\ee
where the logistic function $\sigma(\beta X) = 1/(1+\exp(-\beta X))$ has $\beta=2.4/\sqrt{2}s(X)$ slope. For large uncertainty $s(X)$ this probability is significantly different from zero well outside the interval $[a,b]$.  Thus crisp logical rules for data with Gaussian distribution of errors are equivalent to fuzzy rules with ``soft trapezoid'' membership functions defined by the difference of the two sigmoids, used with crisp input value. The slope of these membership functions, determined by the parameter $\beta$, is inversely proportional to the uncertainty of the inputs.

In the C-MLP2LN neural model  \cite{l-tnn00} such membership functions are computed by the network ``linguistic units''  $L(X;a,b)=\sigma(\beta(X-a))-\sigma(\beta(X-b))$. Relating the slope $\beta$ to the input uncertainty allows to calculate probabilities in agreement with the Monte Carlo sampling. Another way of calculating probabilities, based on the softmax neural outputs $p(C_j|\bX;M) = O_j(\bX)/\sum_{i} O_i(\bX)$ has been presented in \cite{iis00}.

Probabilities $p(C_i|G_\bX;M)$ depend in a continuous way on intervals defining linguistic variables. The error function:

\be
E(M,\bS)=\frac{1}{2}\sum_{\bX}\sum_i \left( p(C_i|G_\bX;M)-\delta(C(\bX),C_i)\right)^2
\ee
depends also on uncertainties of inputs $\bS$. Several variants of such models may be considered, with Gaussian or conical (triangular-shaped) assumptions for input distributions, or neural models with bicentral transfer functions in the first hidden layer. Confusion matrix computed using probabilities instead of the number of yes/no errors allows for optimization of the error function using gradient-based methods. This minimization may be performed directly or may be presented as a neural network problem with special network architecture.

Uncertainties $s_i$ of the values of features may be treated as additional adaptive parameters for optimization. To avoid too many new adaptive parameters optimization of all, or perhaps of a few groups of $s_i$ uncertainties, is replaced by common $\rho$ factors defining the percentage of assumed uncertainty for each group.

This approach leads to the following important improvements for any rule-based system:

\begin{itemize}
\item Crisp logical rules provide basic description of the data, giving maximal comprehensibility.
\item Instead of 0/1 decisions probabilities of classes $p(C_i|\bX;M)$ are obtained.
\item Inexpensive gradient method are used allowing for optimization of very large sets of rules.
\item Uncertainties of inputs $s_i$ provide additional adaptive parameters.
\item Rules with wider classification margins are obtained, overcoming the brittleness problem of some rule-based systems.
\end{itemize}

Wide classification margins are desirable to optimize the placement of decision borders, improving generalization of the system. If the vector $\bX$ of an unknown class is quite typical for one of the classes $C_k$ increasing uncertainties $s_i$ of $X_i$ inputs to a reasonable value (several times the real uncertainty, estimated for a given data) should not decrease the $p(C_k|G_\bX;M)$ probability significantly. If it does the case $\bX$ may be close to the class border and analysis of $p(C_i|G_\bX;\rho, s_i, M)$ as a function of each $s_i$ is needed. These probabilities allow to evaluate the influence of different features on classification. If simple rules are available such explanation may be satisfactory.

Otherwise to gain understanding of the whole data a similarity-based approach to classification and explanation is worth trying. Prototype vectors $\R_i$ are constructed using a clusterization, dendrogram or a decision tree algorithm. Positions of the prototype vectors $\R_i$, parameters of the similarity measures $D(\bX,\bR)$ and other adaptive parameters of the system are then optimized using a general framework for similarity-based methods \cite{SBM}. This approach includes radial basis function networks, clusterization procedures, vector quantization methods and generalized nearest neighbor methods as special examples. An explanation in this case is given by pointing out to the similarity of the new case $\bX$ to one or more of the prototype cases $\R_i$.

Similar result is obtained if the linear discrimination analysis (LDA) is used -- instead of a sharp decision border in the direction perpendicular to LDA hyperplane a soft logistic function is used, corresponding to a neural network with a single neuron. The weights and bias are fixed by the LDA solution, only the slope of the function is optimized.

\section{Real-life example}

Hepatobiliary disorders data, used previously in several studies \cite{Hayaset,Hayasetc,Pal,hayashiexp}, contains medical records of 536 patients admitted to a university affiliated Tokyo-based hospital, with four types of hepatobiliary disorders: alcoholic liver damage (AL), primary hepatoma (PH), liver cirrhosis (LC) and cholelithiasis (CH).
Each record includes results of 9 biochemical tests and a sex of the patient. The same 163 cases as in \cite{hayashiexp} were used as the test data.

In the previous work three fuzzy sets per each input were assigned using recommendation of the medical experts. A fuzzy neural network was constructed and trained until 100\% correct answers were obtained on the training set. The accuracy on the test set varied from less than 60\% to a peak of 75.5\%. Although we quote this result in the Table \ref{tab:hepato} it seems impossible to find good criteria that will predict when the training should be stopped to give the best generalization. Fuzzy rules equivalent to the fuzzy network were derived but their accuracy on the test set was not given. This data has also been analyzed by Mitra {\em et al.} \cite{mitra,Pal} using a knowledge-based fuzzy MLP system with results on the test set in the range from 33\% to 66.3\%, depending on the actual fuzzy model used.

For this dataset classification using crisp rules was not too successful. The initial 49 rules obtained by C-MLP2LN procedure gave 83.5\% on the training and 63.2\% on the test set. Optimization did not improve these results significantly. On the other hand fuzzy rules derived using the FSM network, with Gaussian as well as with triangular functions, gave similar accuracy of 75.6-75.8\%. Fuzzy neural network used over 100 neurons to achieve 75.5\% accuracy, indicating that good decision borders in this case are quite complex and many logical rules will be required. Various results for this dataset are summarized in Table \ref{tab:hepato}.

\begin{table}[htb]
\caption{Results for the hepatobiliary disorders.
Accuracy on the training and test sets, in \%. Top results are achieved eliminating classes or predicting pairs of classes. All calculations are ours except where noted.}\label{tab:hepato}
\begin{center}
\begin{tabular}{|l|r|r|}
\hline
Method 					& Training set & Test set\\

\hline
FSM-50, 2 most prob. classes	& 96.0 		& 92.0 \\
FSM-50, class 2+3 combined 	& 96.0 		& 87.7 \\
FSM-50, class 1+2 combined 	& 95.4 		& 86.5 \\
Neurorule \cite{Hayaset}		& 85.8 		& 85.6 \\
Neurolinear \cite{Hayaset}	& 86.8 		& 84.6 \\

\hline\hline

1-NN, weighted (ASA) 		& 83.4 		& 82.8 \\
FSM, 50 networks			& 94.1	        & 81.0 \\
1-NN, 4 features 			& 76.9 		& 80.4 \\
K* method 				& -- 			& 78.5 \\
kNN, k=1, Manhattan 		& 79.1 		& 77.9 \\
FSM, Gaussian functions 		& 93 			& 75.6 \\
FSM, 60 triangular functions 	& 93 			& 75.8 \\
IB1c (instance-based) 		& -- 			& 76.7 \\
C4.5 decision tree 			& 94.4 		& 75.5 \\
Fuzzy neural network \cite{Pal,mitra}& 100 	& 75.5 \\
Cascade Correlation 		& -- 			& 71.0 \\
MLP with RPROP 			& -- 			& 68.0 \\
Best fuzzy MLP model \cite{Hayasetc}& 75.5 		& 66.3 \\
C4.5 decision rules 			& 64.5 		& 66.3 \\
DLVQ (38 nodes) 			& 100 		& 66.0 \\
LDA (statistical) 			& 68.4 		& 65.0 \\
49 crisp logical rules 		& 83.5 		& 63.2\\
FOIL (inductive logic) 		& 99 		& 60.1 \\
T2 (rules from decision tree)	& 67.5 		& 53.3 \\
1R (rules) 					& 58.4 		& 50.3 \\
Naive Bayes 				&-- 			& 46.6 \\
IB2-IB4 					&81.2-85.5 	& 43.6-44.6 \\
\hline
\end{tabular}
\end{center}
\end{table}

FSM creates about 60 Gaussian or triangular membership functions achieving accuracy of 75.5-75.8\%. Rotation of these functions (i.e. introducing linear combination of inputs to the rules) does not improve this accuracy. We have also made 10-fold crossvalidation tests on the mixed data (training plus test data), achieving similar results. Many methods give rather poor results on this dataset, including various variants of the instance-based learning (IB2-IB4, except for the IB1c, which is specifically designed to work with continuous input data), statistical methods (Bayes, LDA) and pattern recognition methods (LVQ).

The best classification results were obtained with the committee of 50 FSM neural networks \cite{FSM,FSMnew} (in Table 1 shown as FSM-50), reaching 81\%. The $k$-nearest neighbors (kNN) with k=1, Manhattan distance function and selection of features gives 80.4\% accuracy (for details see \cite{duch-fuzzy}) and after feature weighting 82.8\% (the training accuracy of kNN is estimated using the leave-one-out method). K* method based on algorithmic complexity optimization gives 78.5\% on the test set, with other methods giving significantly worse results.

The confusion matrix obtained on the training data from the FSM system, averaged over 5 runs and rounded to integer values is (rows - predicted, columns - required):

\be
\left(\begin{array}{l r r r r}
			&AL  &  PH &   LC&   CH\\
AL   			&70   &    6 &    3 &      3\\
PH   			& 3    &121 &    3 &      1\\
LC    		& 1    &   8 &   77 &      2\\
CH    		& 0    &   0 &    0  &    72
\end{array} \right)
\nonumber
\ee

Looking at the confusion matrix one may notice that the main problem comes from predicting AL or LC when the true class is PH. The number of vectors that are classified incorrectly with high confidence (probability over 0.9) in the training data is 10 and in the test data 7 (only 4.3\%).
Rejection of these cases increases confidence in classification, as shown in Fig. 1.

\begin{figure} [hbt]
\begin{center}
\includegraphics[width=0.4\textwidth]{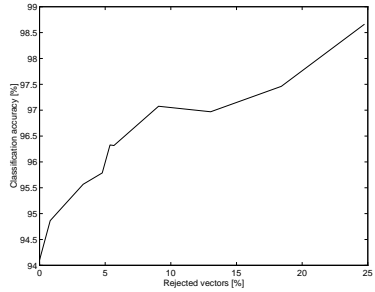}
\end{center}
\caption{Relation between the accuracy of classification and the rejection rate.\label{pstwa}}
\end{figure}

In \cite{Hayaset,Hayasetc} a ``relaxed success criterion'' has been used, counting as a success if the first two strongly excited output neurons contain the correct class. This is equivalent to elimination of 2 classes, leaving the combination of other two as the most probable. In this case accuracy improves, reaching about 90\%. In \cite{Hayaset} two rule extraction methods, {\em Neurorule} and {\em Neurolinear} are used, and the best test set results reach 88.3\% and 90.2\% respectively. Unfortunately true classification accuracy results of these methods are significantly worse then those quoted in Table 1, reaching only 48.4\% ({\em Neurorule} ) and 54.4\% ({\em Neurolinear} ) \cite{Hayaset} on the test set.

We have used here the elimination approach defining first a committee of 50 FSM networks that classify 81\% of cases correctly with high reliability, while cases which cannot be reliably classified are passed to the second stage, in which elimination of pairs of classes (1+2 or 2+3) is made. Training a ``supersystem'', with the error function given by Eq. (\ref{err2}) that tries to obtain the true class as one of the two most probable classes, gives 92\% correct answers on the test and 96\% on the training set. This high accuracy unfortunately drops to 87\% if a threshold of $p\geq 0.2$ is introduced for the second class. In any case reliable  diagnosis of about 80\% of the test cases is possible and for the half of the remaining cases one can eliminate two classes and assign the case under consideration to a mixture of the remaining two classes.

\section{Discussion}

Even when classification in a multi-class problem is poor a useful decision support can still be provided using a classifier that is able to predict some cases with high confidence and an eliminator that can reliably eliminate several classes. The case under consideration most probably belongs to a mixture of remaining classes. Eliminators are build by analysis of confusion matrices and training classifiers with modified error functions.

Since not all classifiers provide probabilities and thus allow to estimate the confidence in their decisions we have described here a universal way to obtain probabilities $p(C_k|\bX;\rho,M)$ using Monte Carlo estimations. Since usually only one new case is evaluated at a time (for example in medical applications) the cost of Monte Carlo simulations is not so relevant. For rule-based systems these probabilities may be determined analytically. Application of these ideas allowed a committee of neural networks to achieve excellent results on medical data that is quite difficult to classify. Further research to determine the best ways to  eliminate some classes and reliably predict mixtures of classes is under way.


\end{document}